\documentclass[10pt,twocolumn,letterpaper]{article}
\usepackage{cvpr}         
\usepackage[accsupp]{axessibility}  

\definecolor{cvprblue}{rgb}{0.21,0.49,0.74}
\definecolor{lightyellow}{RGB}{255, 255, 204}
\definecolor{lightscarlet}{RGB}{255, 204, 204}
\definecolor{lightgreen}{rgb}{0.88,1,0.88}
\usepackage[pagebackref,breaklinks,colorlinks,allcolors=cvprblue]{hyperref}
\usepackage[capitalize]{cleveref}
\usepackage[utf8]{inputenc} 
\usepackage[T1]{fontenc}    
\usepackage{url}            
\usepackage{booktabs}       
\usepackage{multirow}
\usepackage{amsfonts}       
\usepackage{nicefrac}       
\usepackage{microtype}      
\usepackage{xcolor}         
\usepackage{graphicx}   
\usepackage{booktabs}   
\usepackage{wrapfig}    
\usepackage{enumitem} 
\usepackage{minitoc}
\usepackage{mathtools}
\usepackage{amsthm}
\usepackage{caption}
\usepackage[dvipsnames]{xcolor}
\usepackage[table]{xcolor} 
\usepackage{pifont}
\usepackage{amssymb}
\usepackage{overpic}

\def\paperID{****} 
\def\confName{****}
\def\confYear{****}

\title{\method: Motion–Language Alignment for Text-to-Human Motion Generation}

\author{\begin{tabular}{cccccccccccc}\multicolumn{3}{c}{Yannan He \textsuperscript{1,2}} & \multicolumn{3}{c}{Garvita Tiwari \textsuperscript{1,2,3}} & \multicolumn{3}{c}{Xiaohan Zhang \textsuperscript{1,2,3}} & \multicolumn{3}{c}{Pankaj Bora \textsuperscript{1,5}} \end{tabular}\\
\begin{tabular}{ccccccccc}\multicolumn{3}{c}{Tolga Birdal \textsuperscript{4}} & \multicolumn{3}{c}{Jan Eric Lenssen \textsuperscript{3}} &  \multicolumn{3}{c}{Gerard Pons-Moll \textsuperscript{1,2,3}}\end{tabular}\\\\
{\small \textsuperscript{1}University of Tübingen, Germany, \qquad \textsuperscript{2} Tübingen AI Center, Germany}\\
{\small\textsuperscript{3}Max Planck Institute for Informatics, Saarland Informatics Campus, Germany}\\
{\small\textsuperscript{4}Imperial College London, UK,\qquad  \small\textsuperscript{5}Zuse School ELIZA, Germany}\\
{\small \url{https://hynann.github.io/molingo/MoLingo.html}
}
}

\usepackage{bm}
\usepackage{algorithm,algpseudocode}
\algrenewcommand\algorithmicrequire{\textbf{Input:}}
\algrenewcommand\algorithmicensure{\textbf{Output:}}

\definecolor{rowhighlight}{gray}{0.9}
\usepackage{mwe}

\newcommand{\method}[0]{{MoLingo}}

\newcommand{\figref}[1]{Fig.~\ref{#1}}
\newcommand{\secref}[1]{Sec.~\ref{#1}}

\newcommand{\tabref}[1]{Tab.~\ref{#1}}

\DeclarePairedDelimiterX{\infdivx}[2]{(}{)}{%
  #1\;\delimsize\|\;#2%
}

\crefname{equation}{eq.}{eq.}
\Crefname{equation}{Eq.}{Eq.}
\crefname{theorem}{thm.}{thms.}
\Crefname{Theorem}{Thm.}{Thms.}
\crefname{conjecture}{conj.}{conjs.}
\Crefname{Conjecture}{Conj.}{Conjs.}
\crefname{proposition}{prop.}{props.}
\Crefname{proposition}{Prop.}{Props.}
\crefname{definition}{dfn.}{dfn.}
\Crefname{definition}{Dfn.}{Dfn.}
\crefname{remark}{remark}{remark}
\Crefname{Remark}{Remark}{Remark}
\Crefname{algorithm}{Alg.}{Alg.}

\crefname{section}{Sec.}{Secs.}
\Crefname{section}{Sec.}{Secs.}
\crefname{equation}{Eq.}{Eqs.}
\Crefname{equation}{Eq.}{Eqs.}
\crefname{figure}{Fig.}{Figs.}
\Crefname{figure}{Fig.}{Figs.}
\crefname{table}{Tab.}{Tabs.}
\Crefname{table}{Tab.}{Tabs.}
\crefname{thm}{Thm.}{Thms.}
\Crefname{thm}{Thm.}{Thms.}
\crefname{conj}{Conj.}{Conjs.}
\Crefname{conj}{Conj.}{Conjs.}
\crefname{dfn}{Dfn.}{Dfns.}
\crefname{dfn}{Dfn.}{Dfns.}
\crefname{remark}{remark}{remarks}
\Crefname{Remark}{Remark}{Remarks}
\crefname{prop}{Prop.}{Prop.}
\Crefname{prop}{Prop.}{Prop.}
\Crefname{algorithm}{Alg.}{Alg.}
\crefname{appendix}{App.}{apps.}
\Crefname{appendix}{App.}{Apps.}
\crefname{appsec}{appendix}{appendices}
\Crefname{appsec}{Appendix}{Appendices}

\renewcommand{\paragraph}[1]{{\vspace{1mm}\noindent \bf #1.}}

\definecolor{bestrow}{RGB}{230, 245, 208}

\begin{document}
\makeatletter
\let\@oldmaketitle\@maketitle
\renewcommand{\@maketitle}{
	\@oldmaketitle
		\begin{center}
\centering
    \captionsetup{type=figure}
    \vspace{-20pt}
    \includegraphics[width=\textwidth]{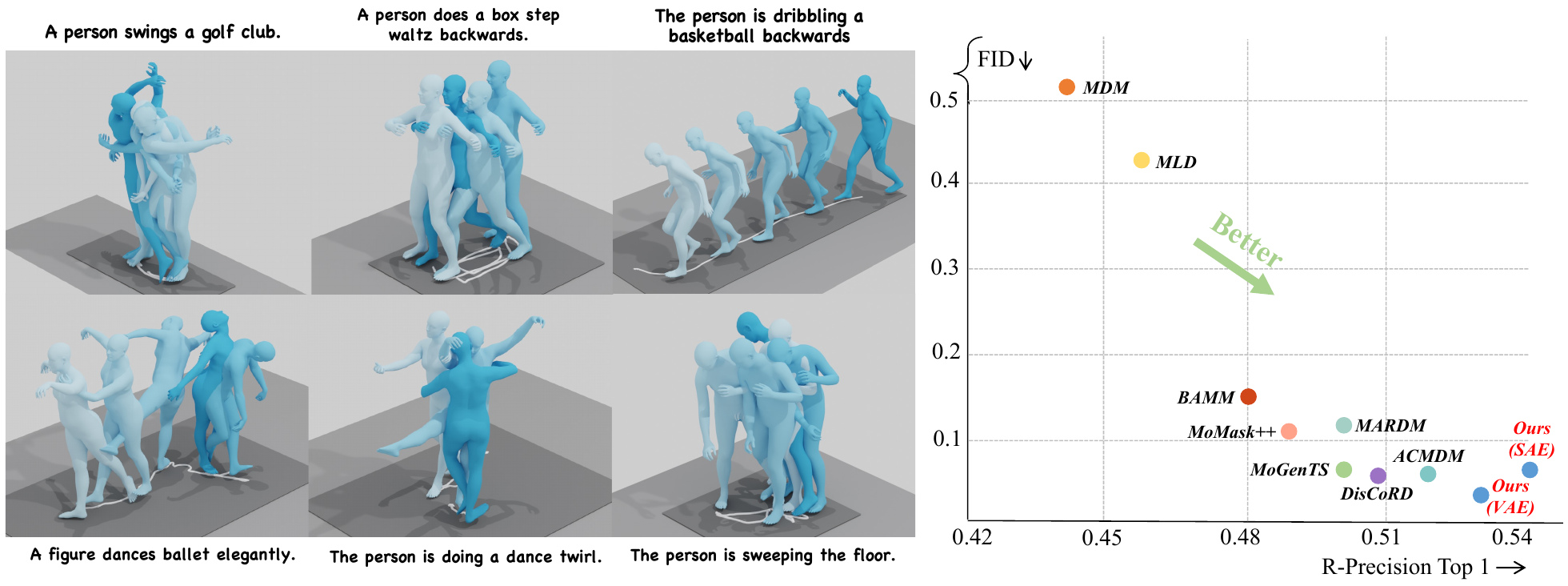}
    	\end{center}
     \vspace{-10pt}
\refstepcounter{figure}\normalfont Figure~\thefigure.  \textit{Left:} Given text prompts, \method{} generates realistic and text-aligned motions, ranging from daily movements like sweeping to more challenging movements like dancing. \textit{Right:} \method{} significantly outperforms previous works in both FID and R-Precision scores. The difference can best be seen in motion, hence we urge the reader to view our project page.
	\label{fig:teaser}
	\newline
}
\makeatother

\maketitle
\begin{abstract}

\vspace{-0.5cm}

We introduce \method, a text-to-motion (T2M) model that generates realistic, lifelike human motion by denoising in a continuous latent space. Recent works perform latent space diffusion, either on the whole latent at once or auto-regressively over multiple latents. In this paper, we study how to make diffusion on continuous motion latents work best. We focus on two questions: (1) how to build a semantically aligned latent space so diffusion becomes more effective, and (2) how to best inject text conditioning so the motion follows the description closely. We propose a semantic-aligned motion encoder trained with frame-level text labels so that latents with similar text meaning stay close, which makes the latent space more diffusion-friendly. We also compare single-token conditioning with a multi-token cross-attention scheme and find that cross-attention gives better motion realism and text–motion alignment. With semantically aligned latents, auto-regressive generation, and cross-attention text conditioning, our model sets a new state-of-the-art in human motion generation on standard metrics and in a user study. We will release our code and models for further research and downstream usage. 

\end{abstract}

\section{Introduction}
\label{sec:intro}

Generating realistic human motion from text is crucial for computer animation, AR/VR entertainment, and building agents that can follow human instructions.
In recent years, text-labeled human motion datasets~\cite{plappert2016kit, Guo2022CVPR} have kicked off this research area, and diffusion-based generative models have brought a major improvement in generated motion quality.
Earlier text-to-motion (T2M) works used diffusion models to denoise motion directly in pose space~\cite{tevet2023human, zhang2022motiondiffuse, zhang2023remodiffuse, camdm, zhou2023emdm, karunratanakul2023gmd, dabral2022mofusion, ghosh2024remos, liang2024intergen, huang2024stablemofusion, li2024unimotion} and achieved good performance. However, denoising raw pose frames is hard because the joint distribution is very complex, and it can also brings artifacts by preserving mocap noise~\cite{amass}.
To address this, recent works first encode motion into a compact latent space~\cite{chen2023executing, motionlcm, motionlcm-v2, zhang2024motion, zhangflashmo}, then run diffusion in that space, and finally decode it back to motion.
Current methods mainly follow two tracks: (1) diffusion on the whole latent space at once~\cite{chen2023executing, motionlcm, motionlcm-v2, jiang2025motionpcm}, and (2) split the sequence into multiple latents and denoise them auto-regressively~\cite{Zhao:DART:2025, meng2024rethinking_marmdm, xiao2025motionstreamer, tuautoregressive, zhu2025motiongpt3}.

In this work, we investigate how to most effectively perform text-to-motion diffusion in a continuous latent space.
There are two key questions here: (1) what makes a good latent space for motion diffusion, and (2) how to inject the text condition most effectively.

For the first question, we study motion latent spaces w.r.t. latent size and semantic alignment, motivated by recent latent space studies for image generation~\cite{yu2025repa, leng2025repae, wu2025representation}. Specifically, we introduce a semantically aligned motion encoder that preserves key temporal structure and pulls atomic motion latents with similar text semantics closer in the latent space. We additionally use frame-level text labels guiding how motion latents are distributed.
Altogether results in a model which can follow the text instructions more faithfully.

For text conditioning, recent works usually either attach a single text token to the motion latents~\cite{tevet2023human, chen2023executing, zhang2023generating, guo2023momask,  pinyoanuntapong2024mmm, pinyoanuntapong2024bamm}, or use that token to modulate the transformer-based diffusion backbone~\cite{meng2024rethinking_marmdm, meng2025absolute}. Our experiments show that a single token is not expressive enough. Instead, we employ multiple text tokens with cross-attention to the motion latents. This results in stronger conditioning. 

Based on these insights, we introduce \method, a masked auto-regressive rectified-flow model that combines semantically aligned latent spaces with multi-token cross-attention. \method\ generates human motion that is both natural and faithful to text instructions, achieving state-of-the-art results in FID, R-Precision, and a user study, including challenging motions such as dancing~\cref{fig:teaser}.

In summary, our \textbf{contributions} are:
\begin{itemize}[noitemsep,leftmargin=*,topsep=0em]
\item We introduce \textbf{\method}, setting the new state-of-the-art in human motion generation using masked auto-regressive rectified flow on a continuous latent space to produce natural and realistic motions.
\item We propose two variants of our model: one with a semantically aligned latent space of text and motion and one with a plain VAE. We show that alignment helps the model to faithfully follow the text instruction. 
\item Instead of conditioning on text based on a single token (common), we find that multiple tokens and cross-attention with motion latents yields significant improvement. 
\end{itemize}
We will make our code and models publicly available for further research and downstream usage.

\section{Related Work}\vspace{-2mm}
\label{sec:related_work}
Human motion is a vastly studied topic. Different from unconditional generation ~\cite{rempe2021humor,yu2025geometric, yu2020character, he2024nrdf}, 
here we review recent advances in text-driven 3D human motion generation, categorized into: human motion generation, evaluation protocols, and representation learning for diffusion models.


\paragraph{Human motion generation}
%
%
Human motions can be parameterized either continuously or discretely. 
For continuous representations, diffusion-based methods generate motions either directly in the pose-frame space~\cite{zhang2022motiondiffuse, tevet2023human, zhang2023remodiffuse, zhang2024finemogen, zhang2025large, zhou2023emdm, dabral2022mofusion, camdm, shafir2023human, karunratanakul2023gmd, liang2024intergen, petrovich2024multi, li2024unimotion, chen2025free, zhang24both, li2026frankenmotion, zhang2024scenic, zhang2024force, zhang2022couch, liang2024omg} or in a learned latent space~\cite{chen2023executing, motionlcm, motionlcm-v2, zhang2024motion, zhangflashmo, meng2024rethinking_marmdm, meng2025absolute, xiao2025motionstreamer, tuautoregressive}.
While diffusion-based approaches can produce diverse and realistic motions, methods that generate in the pose-frame space suffer from the noise present in mocap-based datasets~\cite{amass}, leading to artifacts in generated motions.
Latent space diffusion models alleviate the issue by first compressing motions into a compact latent space. However, encoding an entire sequence into a single latent vector~\cite{chen2023executing, motionlcm, motionlcm-v2, jiang2025motionpcm} can lose important temporal details and fine-grained motion cues. 
%
%
Vector-quantization (VQ)–based next-token prediction methods~\cite{
chuan2022tm2t, zhang2023generating, jiang2024motiongpt, guo2023momask, pinyoanuntapong2024mmm, pinyoanuntapong2024bamm, wan2024tlcontrol, javed2024intermask, wang2025motiondreamer, EgoLM, chen2025language, guo2025snapmogen, pinyoanuntapong2025maskcontrol, ghosh2025duetgen, zhang2025kinmo, liu2025gesturelsm, shan2025mojito} first discretize continuous motions into a series of codebook tokens, then use auto-regressive transformers to predict subsequent token entry conditioned on text, treating motion as a foreign language.
While this discretization preserves more temporal information from the input motion and makes training easier via cross-entropy, mapping continuous motion to a finite codebook still introduces quantization error. This reduces fine-grained details and realism, especially for challenging or fast motions.
%
%
More recently, inspired by advances in image generation~\cite{li2024autoregressive}, several works~\cite{meng2024rethinking_marmdm, xiao2025motionstreamer, zhu2025motiongpt3, tuautoregressive, zhang2025vibes} have adopted continuous-valued latents for diffusion-based auto-regressive motion generation, improving motion quality and reducing the reconstruction artifacts of VQ-based methods. We follow this line, but introduce with a stronger text-conditioning mechanism and a semantically aligned latent space, which together enable state-of-the-art motion generation performance.


\begin{figure}[t]
\vspace{-3mm}
\centering
    \begin{overpic}[trim=0cm 0cm 0cm 0cm,clip, width=\linewidth]{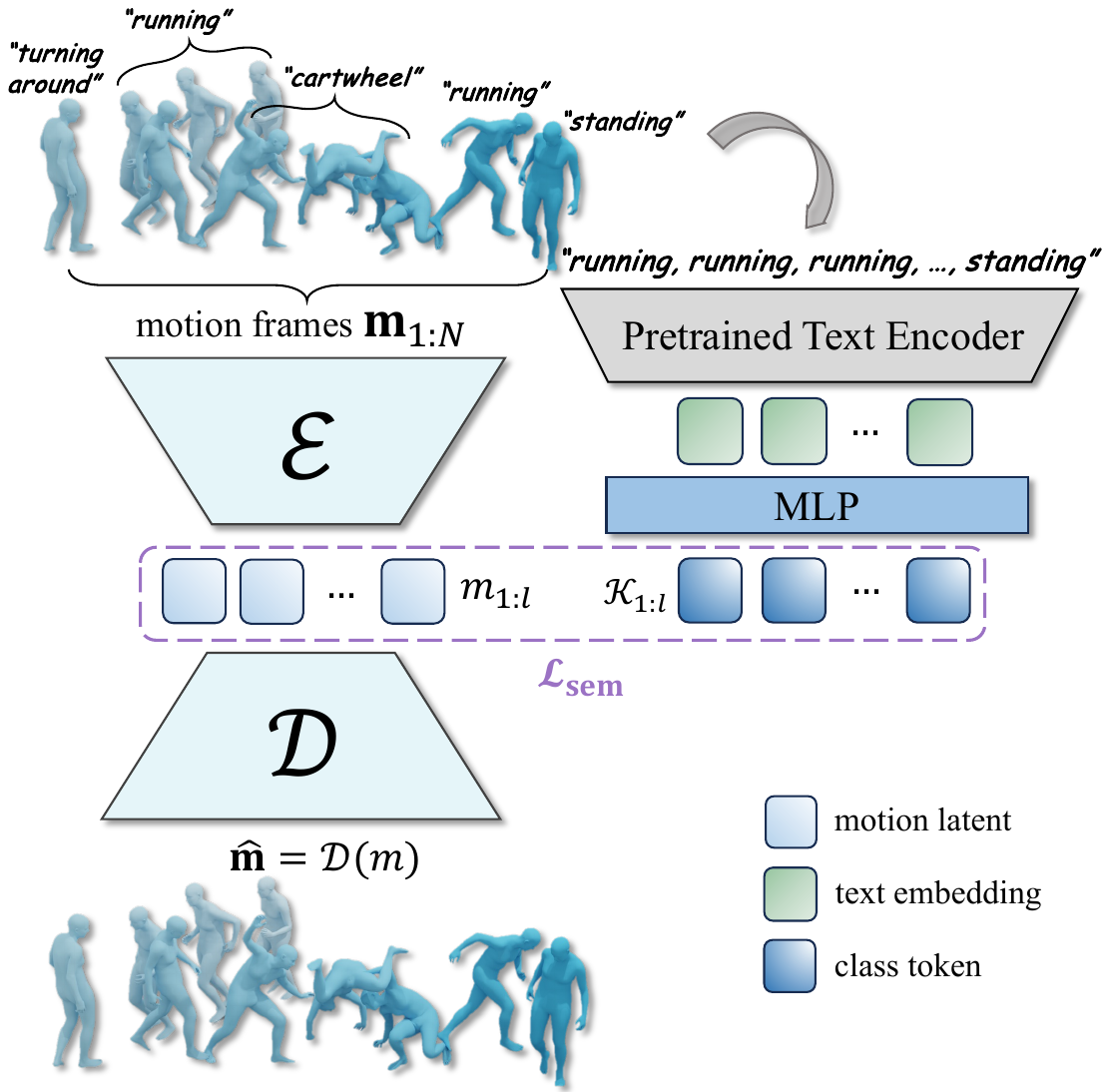}
    \end{overpic}
    \caption{\textbf{Semantically aligned autoencoder architecture.} 
    The model comprises an encoder–decoder autoencoder for motion sequences and a parallel text-encoding branch that maps frame-level text labels into class tokens. A cosine-similarity loss $\mathcal{L}_{\text{sem}}$ is applied to align the motion latents with their corresponding class tokens.}
    \vspace{-3mm}
\label{fig:sae}
 \vspace{-0.5pt}
\end{figure}


\paragraph{Motion parameterization and evaluation protocols}
Human motion is naturally represented as a sequence of pose frames. Both, how we parameterize the pose configuration and how we quantify motion quality, are crucial for assessing realism and semantic faithfulness.
FID~\cite{heusel2017gans} is the most widely used metric to measure realism and diversity for generative models. To better align with human perception, recent works explore directly predicting human preference scores~\cite{voas2023best, motioncritic2025}.
In this context, HumanML3D~\cite{Guo2022CVPR} introduced a 263-dimensional motion representation together with a motion feature extractor for computing metrics such as FID. In addition to rotations, this representation also contains joint positions, velocities, and foot-contact signals, which serve as extra regularization.
This Guo-263 protocol enabled robust comparison for a wide range of text-to-motion works ~\cite{tevet2023human, chen2023executing, zhang2022motiondiffuse, zhang2023generating, guo2023momask}, but subsequent studies have revealed limitations~\cite{petrovich23tmr, li2024unimotion, petrovich2024multi, lbensabath2024, meng2024rethinking_marmdm, xiao2025motionstreamer}.
\textbf{(1)} ~\cite{meng2024rethinking_marmdm} pointed out that the 263D representation is redundant. They keep only the first 67 dimensions, corresponding to the pelvis trajectory and local joint positions, resulting in the MARDM-67 protocol. 
\textbf{(2)} ~\cite{xiao2025motionstreamer} pointed out that the rotation components in Guo-protocol are derived via inverse kinematics (IK), which introduces errors. They rebuilt the representation into 272 dimensions by taking rotations directly from AMASS~\cite{amass}, thereby avoiding IK artifacts.
They further adopt a better-designed cross-modal embedding space, TMR~\cite{petrovich23tmr}, which provides a more semantically coherent latent space for evaluation, resulting in the MS-272 protocol.
Recent works have reported FID using TMR evaluator~\cite{petrovich2024multi, li2024unimotion, xiao2025motionstreamer, guo2025snapmogen}. For the standard 263-dimensional representation, we also employ the TMR evaluator on the original 263D representation, referred to as TMR-263. Together with MARDM-67, we report results across all these evaluators to ensure a fair and comprehensive evaluation.


\paragraph{Representation learning for diffusion models}
Learning effective latent representations is crucial for high-quality generation with latent diffusion models.
Recent works have focused on discovering latent spaces suitable to diffusion, exploring what constitutes a ``diffusable'' representation.
REPA~\cite{yu2025repa} introduces an approach that aligns noisy input states in diffusion models with representations from pretrained visual encoders.
REPA-E~\cite{leng2025repae} proposes to train the diffusion model and its VAE tokenizer end-to-end, achieving stable and effective joint training of both components.
REG~\cite{wu2025representation} proposes to entangle low-level image latents with a single high-level class token from pretrained foundation models, enhancing global control for the denoising process.
~\cite{lee2025latent} integrates masked autoencoders into the diffusion framework, while recent work has also explored semantic regularization losses~\cite{niu2025semantic} to improve latent space structure.
Motivated by these insights, we investigate how to enrich motion latent spaces with semantic information to facilitate the diffusion process and improve generation quality.
\vspace{-0.2cm}


\begin{figure}[t]
\vspace{-3mm}
\centering
    \begin{overpic}[trim=0cm 0cm 0cm 0cm,clip, width=\linewidth]{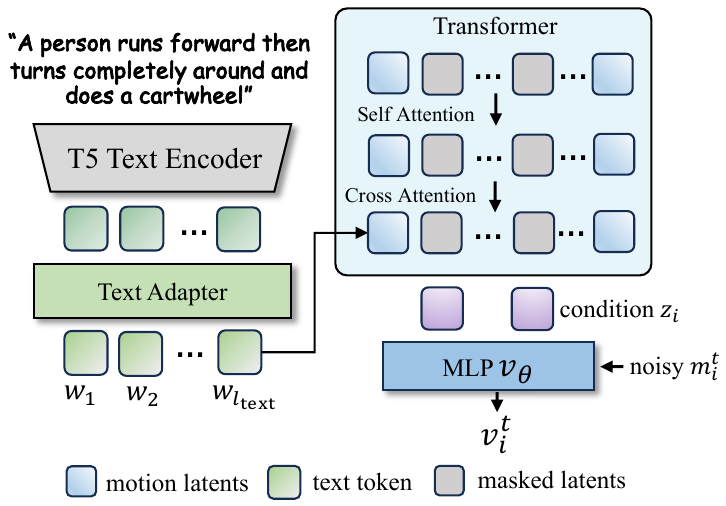}
    \end{overpic}
    \caption{\textbf{Auto-regressive flow-based latent denoising.} Our generation model uses a standard transformer decoder to obtain conditioning vectors $z$, which guides an MLP in iteratively refining latents. During training the motion latents are randomly masked and replaced with learnable tokens. During inference, we initialize with \textbf{fully masked latents}, iteratively denoise them, and decode the final latents to obtain the generated motion.}
    \vspace{-3mm}
\label{fig:ar}
 \vspace{-0.5pt}
\end{figure}

\section{Method} 
In this section, we introduce our method for realistic human motion generation. We start by giving an overview of the solved task and our method.

\paragraph{Human motion generation task}
We aim to generate a sequence $\mathbf{m}_{1:N}$ of 3D human motion with length $N$, where each $\mathbf{m}_i \in \mathbb{R}^{D}$ represents a human pose configuration.  The generation process is conditioned on a textual description $c$.  

\paragraph{Method overview}
As illustrated in~\figref{fig:sae} and \figref{fig:ar}, our approach consists of two key components: $(1)$ a motion encoder that maps motion sequences to semantically aligned continuous-valued motion latents (\secref{sec:tokenizer}) and $(2)$ a masked auto-regressive transformer with rectified flow~\cite{liu2022flow} output heads, progressively denoising these latents (\secref{sec:model}).
Similar to ~\cite{meng2024rethinking_marmdm, xiao2025motionstreamer}, we perform auto-regressive generation directly in the continuous domain using a flow-based objective, which avoids poor reconstruction performance observed in earlier methods~\cite{zhang2023generating, jiang2024motiongpt, guo2023momask, pinyoanuntapong2024mmm, pinyoanuntapong2024bamm}.


\begin{table*}[th]
\begin{center}
\centering
\caption{\textbf{Quantitative results on the MARDM-67 evaluator.}
We compare our method with a broad set of motion generation approaches, from early models~\cite{tevet2023human, chen2023executing} to recent ones~\cite{meng2025absolute, guo2025snapmogen}, covering pose-frame diffusion~\cite{tevet2023human}, single-vector latent diffusion~\cite{chen2023executing, motionlcm-v2}, VQ-based next-token prediction~\cite{guo2023momask, pinyoanuntapong2024mmm, pinyoanuntapong2024bamm, yuan2024mogents, cho2024discord, guo2025snapmogen}, and continuous-valued auto-regressive models~\cite{meng2024rethinking_marmdm, meng2025absolute}. We report the mean results over 20 independent runs, and the $\pm$ values indicate the $95\%$ confidence interval. Our method achieves state-of-the-art FID, R-Precision, and CLIP-Score. Green cells highlight the best scores, and yellow cells the second best.\vspace{-0.2cm}
}
\resizebox{0.9\textwidth}{!}{
\begin{tabular}{lcccccc}
\toprule
\multirow{2}{*}{Methods} &
\multirow{2}{*}{FID\,$\downarrow$} &
\multicolumn{3}{c}{R-Precision} &
\multirow{2}{*}{CLIP-Score\,$\uparrow$} &
\multirow{2}{*}{MModality\,$\uparrow$} \\
\cline{3-5}
 & & Top 1\ $\uparrow$ & Top 2\ $\uparrow$ & Top 3\ $\uparrow$ & & \\
\hline
Real  & $0.000^{\pm.000}$ & $0.503^{\pm.002}$ & $0.696^{\pm.001}$ & $0.795^{\pm.002}$ & $0.639^{\pm.001}$ & - \\
\midrule
MDM-50Step~\cite{tevet2023human} & $0.518^{\pm.032}$ & $0.440^{\pm.007}$ & $0.636^{\pm.006}$ & $0.742^{\pm.004}$ & $0.578^{\pm.003}$ & \cellcolor{lightgreen}$3.604^{\pm.031}$   \\ 
MLD~\cite{chen2023executing} & $0.431^{\pm.014}$ & $0.461^{\pm.004}$ & $0.651^{\pm.004}$ & $0.750^{\pm.003}$ & $0.610^{\pm.003}$ & \cellcolor{lightyellow}$3.506^{\pm.031}$  \\ 
MoMask~\cite{guo2023momask} & $0.116^{\pm.006}$ & $0.490^{\pm.004}$ & $0.687^{\pm.003}$ & $0.786^{\pm.003}$ & $0.637^{\pm.003}$ &  $1.309^{\pm.058}$  \\
MMM~\cite{pinyoanuntapong2024mmm} & $0.132^{\pm.004}$ & $0.487^{\pm.003}$ & $0.683^{\pm.002}$ & $0.782^{\pm.001}$ & $0.635^{\pm.003}$ & $1.455^{\pm.106}$  \\
BAMM~\cite{pinyoanuntapong2024bamm} & $0.147^{\pm.005}$ & $0.480^{\pm.003}$ & $0.671^{\pm.003}$ & $0.772^{\pm.003}$ & $0.630^{\pm.001}$ & $1.954^{\pm.070}$  \\
MogenTS~\cite{yuan2024mogents} & $0.060^{\pm.004}$  & $0.500^{\pm.003}$ & $0.694^{\pm.002}$ & $0.793^{\pm.002}$ & $0.636^{\pm.001}$ & $0.882^{\pm.032}$  \\
MLD++~\cite{motionlcm-v2} & $2.027^{\pm.021}$  & $0.500^{\pm.003}$ & $0.691^{\pm.002}$ & $0.789^{\pm.001}$ & $0.639^{\pm.002}$ & $1.924^{\pm.065}$\\
MotionLCM-V2~\cite{motionlcm-v2} & $2.267^{\pm.023}$  & $0.501^{\pm.002}$ & $0.693^{\pm.002}$ & $0.790^{\pm.002}$ & $0.640^{\pm.003}$ & $1.780^{\pm.062}$  \\ 
DisCoRD~\cite{cho2024discord} & \cellcolor{lightyellow}$0.053^{\pm.004}$ & $0.506^{\pm.003}$ & $0.699^{\pm.003}$ & $0.796^{\pm.003}$ & $0.645^{\pm.001}$  & $1.303^{\pm.047}$ \\ 
MARDM-$\mathbf{v}$~\cite{meng2024rethinking_marmdm} & $0.114^{\pm.007}$ & $0.500^{\pm.004}$ & $0.695^{\pm.003}$ & $0.795^{\pm.003}$ & $0.642^{\pm.002}$ & $2.231^{\pm.071}$  \\ 
ACMDM-S-PS22~\cite{meng2025absolute} & $0.109^{\pm.005}$ & $0.508^{\pm.002}$ & $0.701^{\pm.003}$ & $0.798^{\pm.003}$ & $0.642^{\pm.001}$ &  $2.156^{\pm.061}$ \\
ACMDM-XL-PS2~\cite{meng2025absolute} & $0.058^{\pm.004}$ &$0.522^{\pm.002}$ & $0.713^{\pm.003}$ & $0.807^{\pm.002}$ & $0.652^{\pm.001}$ & $2.077^{\pm.083}$  \\
MoMask++~\cite{guo2025snapmogen} & $0.108^{\pm.007}$ & $0.492^{\pm.003}$ & $0.683^{\pm.003}$ & $0.782^{\pm.002}$ & $0.634^{\pm.001}$ &  $1.259^{\pm.054}$  \\
\midrule
MoLingo (VAE)  & \cellcolor{lightgreen}$0.049^{\pm{.003}}$ & \cellcolor{lightyellow}$0.528^{\pm{.002}}$ & \cellcolor{lightyellow}$0.721^{\pm{.002}}$ & \cellcolor{lightyellow}$0.815^{\pm{.002}}$ & \cellcolor{lightyellow}$0.672^{\pm{.001}}$ & $1.414^{\pm.049}$  \\
MoLingo (SAE)  & $0.066^{\pm{.003}}$ & \cellcolor{lightgreen}$0.544^{\pm{.002}}$ & \cellcolor{lightgreen}$0.739^{\pm{.002}}$ & \cellcolor{lightgreen}$0.832^{\pm{.002}}$ & \cellcolor{lightgreen}$0.686^{\pm{.001}}$ & $1.226^{\pm.063}$  \\
\bottomrule
\end{tabular}}
\vspace{-0.2cm}
\label{tab:comp_guo_67}
\vspace{-0.3cm}
\end{center}
\end{table*}


\vspace{-1mm}\subsection{Encoding Motions into Latents}\vspace{-1mm}
\label{sec:tokenizer}
We consider three variants of motion autoencoders: a variational autoencoder (VAE), a vanilla autoencoder (AE), and we further propose a semantically aligned autoencoder (SAE). 
Since our encoder–decoder is built with 1D temporal convolutions, the sequential structure of the input motion is preserved by encoding $N$ motion frames into $l = N/h$ latents, resulting in motion latents \mbox{$m_{1:l} \in \mathbb{R}^{l\times d}$}, where $d$ is the latent dimension. In practice, we adopt a causal architecture~\cite{xiao2025motionstreamer} where both the motion encoder and decoder consist of stacked 1D convolutional layers.

\paragraph{Semantically aligned autoencoder (SAE)}
We introduce a continuous-valued autoencoder that preserves fine-grained human motion details while maintaining semantic for each latent, as shown in \cref{fig:sae}.
To obtain a more structured motion latent space with rich semantics, we encourage motion latents corresponding to similar semantics to be close in the latent space.
In practice, we use the BABEL dataset~\cite{BABEL:CVPR:2021}, which provides frame-level textual label annotations for each motion sequence.
For each motion latent $m_j$, we collect the text labels from the motion frames temporally aligned with that latent, encode them using a frozen text encoder, average the resulting embeddings, and apply a linear projection to match the motion latent dimension, getting the class token $\kappa_j$.
We then treat these class tokens as the teacher to guide the motion latent distribution. 
For each training batch, we apply a cosine similarity loss to encourage closer alignment between motion latents and their corresponding class tokens.
However, since BABEL contains highly repetitive frame-level annotations, consecutive motion latents may be assigned an identical class token, leading to overly strong or incorrect alignment. To address this, we first compute a similarity score between consecutive class tokens: $\Delta_i = \langle \kappa_i, \kappa_{i+1} \rangle, 1 \leq i < B, B=b\times l$, and discard class tokens and their corresponding motion latents whose $\Delta_i$ exceeds a threshold $\tau$, resulting in the filtered index set $\mathcal{I}$. Here $\langle \cdot, \cdot \rangle$ denotes the cosine similarity, and $b$ is the batch size.
We then minimize the semantic loss from the text-induced distribution to the motion-induced one:
\begin{equation}
\mathcal{L}_{\text{sem}} = \frac{1}{|\mathcal{I}|} \sum_{i\in \mathcal{I}} \left(1 - \frac{m_i \cdot \kappa_i}{\|m_i\| \|\kappa_i\|}\right)
\end{equation}

\begin{table*}[t]
\centering
\caption{\textbf{Ablation studies.} We analyze: (1) different text-conditioning mechanisms, and (2) the effect of using a semantically aligned latent space. \textbf{AdaLN} denotes single-text-token conditioning with DiT-style modulation; note that the first row corresponds to MARDM~\cite{meng2024rethinking_marmdm}.
 \textbf{CrossAttn} denotes multi-token cross-attention conditioning. We conduct all experiments using the MARDM-67 evaluator with 4× downsampling and latent dimensions of $16$. We adopt adapter depth $6$ for T5+CrossAttn setting.}
\setlength{\tabcolsep}{6pt}
\resizebox{0.9\textwidth}{!}{
\begin{tabular}{c|c|c|ccccc}
\toprule
Conditioning &
Text &
Autoencoder &
\multirow{2}{*}{FID $\downarrow$} &
\multicolumn{3}{c}{R-Precision} &
\multirow{2}{*}{CLIP-Score  $\uparrow$}\\
\cline{5-7}
Mechanism & Encoder & Type &  & Top 1 $\uparrow$ & Top 2 $\uparrow$ & Top 3 $\uparrow$ \\
\hline
\multirow{2}{*}{AdaLN} & CLIP & \multirow{2}{*}{AE} & $0.114^{\pm{.007}}$ & $0.500^{\pm{.004}}$ & $0.695^{\pm{.003}}$ & $0.795^{\pm{.003}}$ & $0.642^{\pm{.002}}$ \\
& T5 & &  $0.077^{\pm{.004}}$ & $0.508^{\pm{.003}}$ & $0.700^{\pm{.003}}$ & $0.795^{\pm{.002}}$ & $0.650^{\pm{.001}}$ \\
\hline
\multirow{3}{*}{CrossAttn} &
\multirow{3}{*}{T5} & VAE & \cellcolor{lightgreen}$0.049^{\pm{.003}}$ & $0.528^{\pm{.002}}$ & $0.721^{\pm{.002}}$ & $0.815^{\pm{.002}}$ & $0.672^{\pm{.001}}$ \\
&  & AE & \cellcolor{lightyellow}$0.051^{\pm{.002}}$ & \cellcolor{lightyellow}$0.533^{\pm{.003}}$ & \cellcolor{lightyellow}$0.727^{\pm{.003}}$ & \cellcolor{lightyellow}$0.820^{\pm{.002}}$ & \cellcolor{lightyellow}$0.673^{\pm{.001}}$  \\
&  & SAE & $0.066^{\pm{.003}}$ & \cellcolor{lightgreen}$0.544^{\pm{.002}}$ & \cellcolor{lightgreen}$0.739^{\pm{.002}}$ & \cellcolor{lightgreen}$0.832^{\pm{.002}}$ & \cellcolor{lightgreen}$0.686^{\pm{.001}}$ \\
\bottomrule
\end{tabular}}
\label{tab:abl_net}
\end{table*}


\paragraph{Training objective}
First, the SAE is trained to preserve motion reconstruction quality via a reconstruction loss:
\begin{equation}
    \mathcal{L}_{\text{recon}} =  \mathcal{L}_{\text{feat}} + \lambda_{\text{joint}} \mathcal{L}_{\text{joint}} + \lambda_{\text{vel}} \mathcal{L}_{\text{vel}}
\end{equation}
where,
\begin{equation}
   \mathcal{L}_{\text{feat}} = \|\mathbf{m} - \mathcal{D}(m)\|_2^2\ \ \ \mathcal{L}_{\text{joint}} = \|\mathcal{T}(\mathbf{m})- \mathcal{T}(\mathcal{D}(m))\|_2^2
\end{equation}
\begin{align}
\mathcal{L}_{\text{vel}} = \frac{1}{N-1} \sum_{n=1}^{N-1} \Big\| &\left(\mathcal{T}(\mathbf{m})_{n+1} - \mathcal{T}(\mathbf{m})_n\right)  \\
&- \left(\mathcal{T}(\mathcal{D}(m))_{n+1} - \mathcal{T}(\mathcal{D}(m))_n\right) \Big\|_2^2\nonumber
\end{align}
where $\mathcal{T}$ represents the function that transforms motion representation to body joint positions. These terms respectively penalize differences in motion representations, joint positions and joint velocities. The final loss is:
\begin{equation}
\mathcal{L}_{\text{SAE}} = \mathcal{L}_{\text{recon}} + \lambda_{\text{sem}} \mathcal{L}_{\text{sem}} + \lambda_{\text{KL}}\mathcal{L}_{\text{KL}}
\end{equation} 
with $\mathcal{L}_{\text{KL}} = D_{\text{KL}}(q(m|\mathbf{m})\ \| \ p(m))$ is the KL Divergence. Please refer to the appendix for more details.


\vspace{-1mm}\subsection{Masked Auto-regressive Motion Generation}\vspace{-1mm}
\label{sec:model}
\paragraph{Input preparation} 
As shown in \figref{fig:ar}, after encoding the input motions into latents $m_{1:l}$, we first encode the prompt $c$ using a frozen T5-Large~\cite{raffel2020exploring} text encoder. To further enhance cross-modal interaction for the subsequent cross-attention, we pass the text embeddings through a text adapter consisting of $l_{\text{adapter}}$ transformer encoder blocks, obtaining the final text representation $\mathbf{w} = \{ w_1, w_2, \dots, w_{l_{\text{text}}} \}$. We then randomly select a subset of motion latents in $m_{1:l}$, and replace them with a learnable mask latent during training. We continue to denote the masked latent sequence as $m_{1:l}$ for simplicity. $\mathbf{w}$ and $m_{1:l}$ then serve as the inputs to the subsequent auto-regressive denoising stage.

\paragraph{Auto-regressive flow-based latent denoising} 
Given the masked sequence of latents $m_{1:l}$, we decompose the joint distribution over all latents using the chain rule of probability:
\begin{equation}
	p(m_1, \dots, m_l) = \prod_{i=1}^l p(m_i\ |\ c, m_1, \dots, m_{i-1}) \textnormal{,}
\end{equation}
and generate a motion sequence by iteratively sampling latents followed by decoding with $\mathcal{D}$.
Each step uses a transformer-based network to sample from the conditional probability $p(m_i\ |\ c, m_1, \dots, m_{i-1})$ walking through a reverse ODE trajectory. In practice, we use rectified flow~\cite{liu2022flow} as our flow-based model. To combine the strengths of auto-regression and rectified flow, following~\cite{li2024autoregressive, fan2024fluid}, we use a standard decoder-only transformer model $\Phi$ to obtain a conditioning vector \mbox{$z_i = \Phi(w_1, \dots, w_{l_{\text{text}}}; m_1, \dots, m_{i-1})$}. Each transformer block consists of self-attention, cross-attention and MLP layers. The self attention and MLP layer operate only on the motion latents $m_{1:l}$, for cross attention, output of the self attention serve as the queries and text tokens $\mathbf{w}$ as keys and values. Then we feed $z_i$ into an MLP $v_{\theta}$ to approximate the reverse distribution:
\begin{equation}
	\mathcal{L}(z_i, m_i) = \mathbb{E}_{m_i,\epsilon,t}[\| v_{\theta}(m_i^t,t, z_i)  - (\epsilon-m_i) \|^2] \textnormal{,}
\end{equation}
where the noisy $m_i^t$ is obtained by linearly interpolating between the Gaussian noise $\epsilon$ and the clean latent: $m_i^t = \alpha_t m_i + \sigma_t \epsilon, \epsilon \sim \mathcal{N}(0, \mathbf{I})$ with $\alpha_t=1-t$ a decreasing and $\sigma_t=t$ an increasing function of time $t$.
During training, we use bidirectional attention so that masked latents can attend to all unmasked latents, and unmasked latents can also see each other.

\paragraph{Inference} 
We initialize all motion latents with the learned mask latent at the start of inference. We then iteratively select one or a subset of latent positions, run the transformer once to obtain the corresponding conditioning vector(s) $z$, and feed $z$ into the MLP to perform the reverse flow-based sampling at those positions, replacing the masked latents with clean ones. This process is repeated until all motion latents are clean. We finally pass the clean latent sequence to the motion decoder $\mathcal{D}$  to produce the generated motion.


\begin{figure*}
  \centering
    \includegraphics[width=\linewidth]{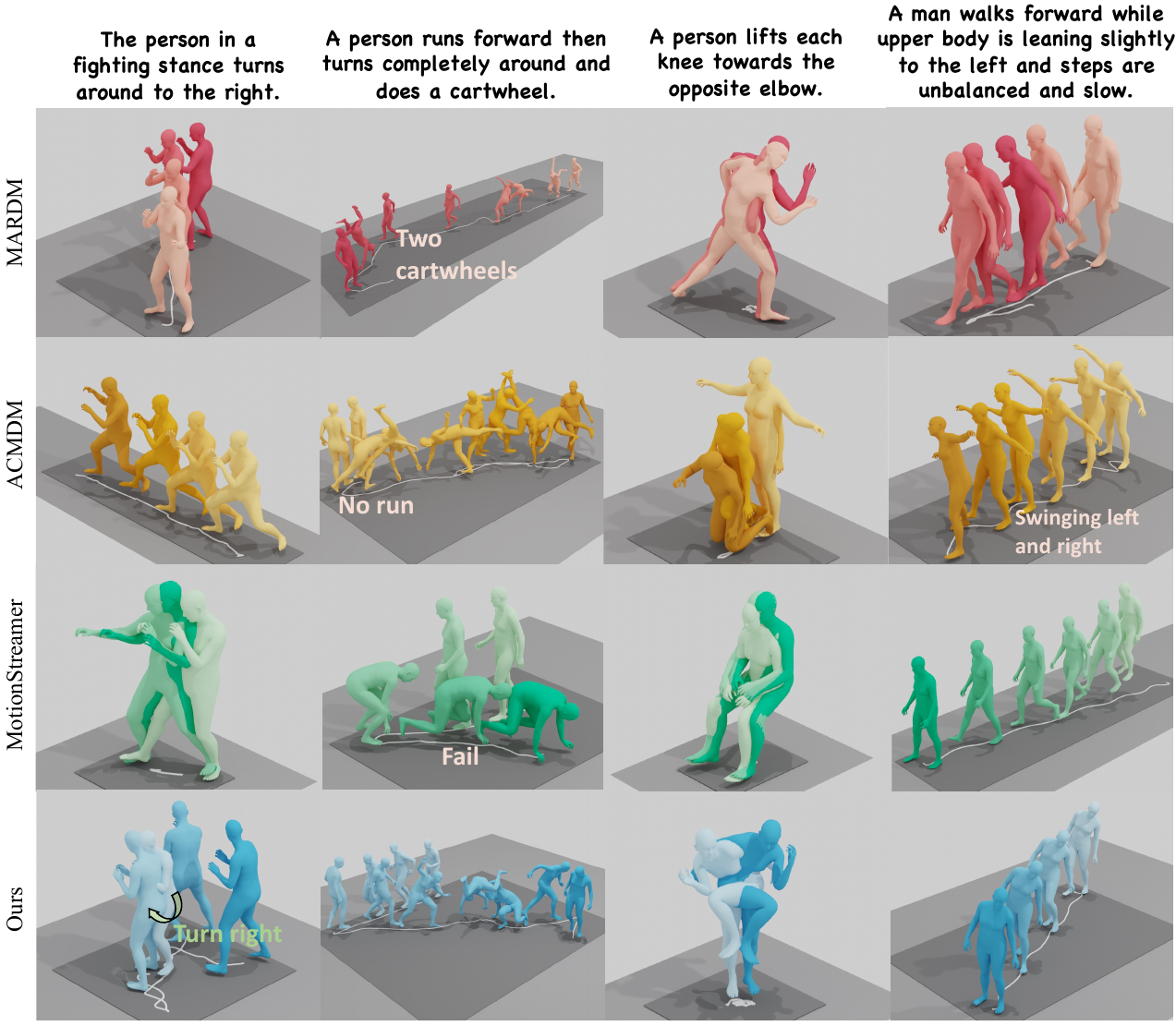}
    \caption{\textbf{Qualitative comparisons} with MARDM~\cite{meng2024rethinking_marmdm}, ACMDM~\cite{meng2025absolute}, and MotionStreamer~\cite{xiao2025motionstreamer}. The color transitions from light to dark to indicate temporal flow. Our method generates more natural and text-aligned motions, ranging from everyday movements to challenging scenarios like cartwheel. In contrast, other methods either fail to follow fine-grained text instructions (e.g., \emph{run first, then turn around, then do a cartwheel}, or \emph{leaning left while walking}), or collapse completely (e.g., knees touching elbows). Our method follows the specified action order while preserving realism. For methods using the MARDM-67 representation~\cite{meng2024rethinking_marmdm,meng2025absolute}, we employ SimpLify~\cite{bogo2016keep} to convert their outputs to SMPL~\cite{smpl} meshes. We refer readers to the project page for more detailed visualizations of dynamic motions.}
    \vspace{-0.5cm}
  \label{fig:gallery}
\end{figure*}

\vspace{-0.1cm}
\section{Experiments}
\label{sec:exp}

In this section, we first present the experimental setup in \secref{sec:exp_setup}, followed by ablation studies in \secref{sec:exp_abl}, qualitative and quantitative comparisons with state-of-the-art methods in \secref{sec:exp_comp}. To better reflect human perception, we also conduct user studies, reported in \secref{sec:exp_comp}.

\subsection{Setup}
\label{sec:exp_setup}

\paragraph{Implementation details}
Our models are implemented in PyTorch. 
To ensure a fair comparison with state-of-the-art human motion generation methods, we use the HumanML3D~\cite{Guo2022CVPR} dataset for training and evaluation. HumanML3D sources motions from AMASS~\cite{amass} and HumanAct12~\cite{guo2020action2motion} and provides two to three manual text annotations per motion. The full dataset consists of 29,024 motions and 87,834 text descriptions. All motions are unified to 20 FPS and last up to 10 seconds. We further use the 272-dimensional 30 FPS version of HumanML3D to train the model for comparison with~\cite{xiao2025motionstreamer}. For SAE training, we use the intersection between BABEL~\cite{BABEL:CVPR:2021} and HumanML3D to apply the semantic loss, and then use the full dataset to apply the remaining losses.
We train the motion autoencoders with a batch size of $256$ for $5000$ epochs, using a learning rate of $5\times10^{-5}$. 
For our auto-regressive latent rectified flow model, we train for about $800$ epochs with a batch size of $256$. The learning rate warms up linearly for the first $100$ epochs and then remains constant at $8\times10^{-4}$. We apply Exponential Moving Average (EMA) to stabilize training. The model is trained with classifier-free guidance (CFG) by replacing $10\%$ of the text prompts into the same null prompt, and at inference time we sample latents with a CFG scale of $5.5$.
The entire training process for our best model configuration in~\cref{tab:comp_guo_67} takes approximately 10 hours with four Nvidia H100 GPUs. Please refer to the supplementary document for further details such as hyperparameter setting, inference speed, etc.


\begin{table*}[t]
\centering
\caption{\textbf{Performance under different SAE configurations.} We ablate three design choices for SAE: (1) the semantic regularization loss (cosine similarity vs. InfoNCE), (2) whether KL divergence is applied jointly, and (3) the weight of $\mathcal{L}_{\text{sem}}$. The configuration with cosine similarity, KL divergence, and a relatively small weight ($0.001$) gives the best SAE design, significantly improving text–motion alignment (R-Precision and CLIP-score) while maintaining FID comparable to SOTA models.}
\resizebox{0.85\textwidth}{!}{
\begin{tabular}{c|c|c|ccccc}
\toprule
semantic & 
\multirow{2}{*}{KL} &
\multirow{2}{*}{$\lambda_{\text{sem}}$} &
\multirow{2}{*}{FID\,$\downarrow$} &
\multicolumn{3}{c}{R-Precision\,$\uparrow$} &
\multirow{2}{*}{CLIP-Score\,$\uparrow$} \\
\cline{5-7}
regularization & & & & Top 1\, & Top 2\, & Top 3\ \\
\hline
\multirow{6}{*}{Cosine} & \ding{55} & $0.1$ & $0.083^{\pm{.003}}$ & $0.527^{\pm{.002}}$ & $0.719^{\pm{.002}}$ & $0.817^{\pm{.002}}$ &  
$0.671^{\pm{.001}}$ \\
 & \checkmark & $0.1$ & $0.071^{\pm{.003}}$ & $0.518^{\pm{.002}}$ & $0.712^{\pm{.002}}$ & $0.808^{\pm{.002}}$ &  
$0.667^{\pm{.001}}$ \\
  & \ding{55} & $0.01$ & $0.081^{\pm{.003}}$ & $0.529^{\pm{.002}}$ & $0.720^{\pm{.002}}$ & $0.822^{\pm{.002}}$ &  
$0.672^{\pm{.001}}$ \\
   & \checkmark & $0.01$ & $0.074^{\pm{.003}}$ & $0.524^{\pm{.002}}$ & $0.722^{\pm{.002}}$ & $0.819^{\pm{.002}}$ &  
$0.673^{\pm{.001}}$ \\
    & \ding{55} & $0.001$ & \cellcolor{lightgreen}$0.062^{\pm{.003}}$ & \cellcolor{lightyellow}$0.535^{\pm{.002}}$ & \cellcolor{lightyellow}$0.731^{\pm{.002}}$ & \cellcolor{lightyellow}$0.826^{\pm{.002}}$ &  
 \cellcolor{lightyellow}$0.680^{\pm{.001}}$ \\
 & \checkmark & $0.001$ & \cellcolor{lightyellow}$0.066^{\pm{.003}}$ & \cellcolor{lightgreen}$0.544^{\pm{.002}}$ & \cellcolor{lightgreen}$0.739^{\pm{.002}}$ & \cellcolor{lightgreen}$0.832^{\pm{.002}}$ & \cellcolor{lightgreen}$0.686^{\pm{.001}}$ \\
\hline
InfoNCE& \ding{55} & $0.001$ & $0.129^{\pm{.007}}$ & $0.523^{\pm{.002}}$ & $0.721^{\pm{.002}}$ & $0.813^{\pm{.002}}$ & $0.671^{\pm{.001}}$  \\ 

\bottomrule
\end{tabular}}
\vspace{-0.2cm}
\label{tab:abl_sae}
\vspace{-0.3cm}
\end{table*}

\paragraph{Evaluation protocol}
To enable a more comprehensive evaluation, we report metrics using MARDM-67 as our main benchmark, which enables us to compare with the most broad range of works, including~\cite{meng2024rethinking_marmdm, meng2025absolute}. Since some other works built on other representations (e.g. 272D), for a fair comparison with methods built on alternative motion representations, we additionally evaluate under their corresponding protocol MS-272. Furthermore, to piorneering a more robust evaluation benchmark for the future research, we for the first time give a most number of previous works and our work on TMR-263 evaluator.
Specifically, we report a set of metrics under different evaluators. We use \textbf{FID} to measure the distance between the generated and real motion distributions, \textbf{R-Precision} and \textbf{CLIP Score} to assess text–motion alignment, and \textbf{MModality} to quantify the diversity of motions generated from the same textual description.


\subsection{Ablation Studies}
\label{sec:exp_abl}

\begin{figure}[t]
\vspace{-3mm}
\centering
    \begin{overpic}[trim=0cm 0cm 0cm 0cm,clip, width=\linewidth]{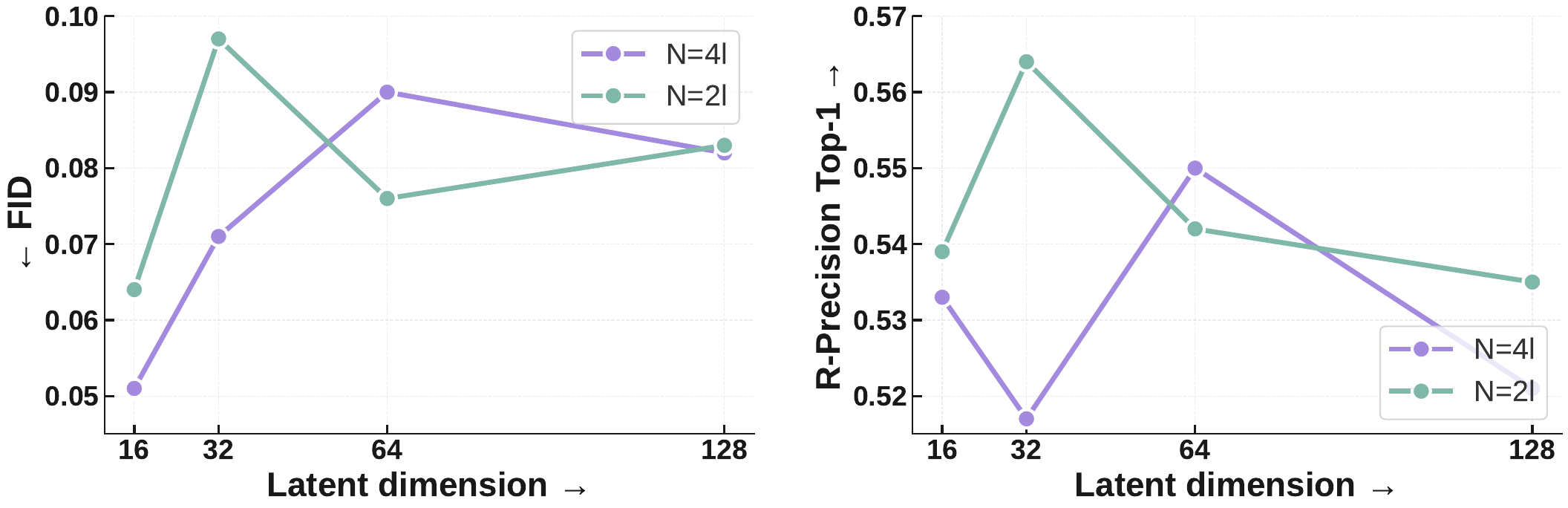}
    \end{overpic}
    \caption{\textbf{Effect of latent dimension and temporal downsampling.} We vary the latent dimension ($16$–$128$) under two temporal compression settings: (4×) and (2×). Overall, the 4× setting gives comparable or better performance than 2×, showing that is beneficial that a single latent encodes 4 frame, even 2× preserves more fine-grained temporal information.
    }
    \vspace{-3mm}
	\label{fig:latent_dim}
 \vspace{-0.5pt}
\end{figure}

We provide a comprehensive ablation on different text-conditioning mechanisms, autoencoder variants followed by an analysis of the effects of different latent sizes in \figref{fig:latent_dim}.

\paragraph{Text condition mechanism} 
\tabref{tab:abl_net} uses MARDM~\cite{meng2024rethinking_marmdm} (row 1) as the baseline to demonstrate the effectiveness of our design choices. We observe that using T5 achieves better performance even without multi-text-token cross attention (row 2), indicating the text encoder plays a key role in preserving rich semantics.
Our multi-token cross attention design achieves clearly better text-motion alignment while achieving lower FID, suggesting that leveraging the full set of text tokens is more expressive.
%


\paragraph{Effect of SAE} 
As shown in \tabref{tab:abl_net}, the proposed SAE clearly outperforms both the vanilla AE and the VAE.
It consistently achieves the best text–motion alignment scores while maintaining a comparable FID to SOTA models, indicating that enforcing semantic alignment yields a more semantically rich latent space that benefits text–motion alignment.
We further investigate different forms of semantic regularization and the influence of the weighting factor $\lambda_{\text{sem}}$ in \tabref{tab:abl_sae}. The variant that incorporates KL-divergence-based supervision with a small weighting factor achieves the best performance. 
For the semantic regularization strategy, we also compare against InfoNCE supervision, a common choice in contrastive learning~\cite{petrovich23tmr}, but it gives worse results.
InfoNCE imposes a hard contrastive constraint that not only pulls together motion latents sharing the same label, but also explicitly pushes apart latents associated with different labels. However, as discussed in \secref{sec:tokenizer}, the available text labels lack diversity: semantically similar motions might all be mapped to the same label.
However, human motion is continuous and ambiguous, and a motion latent corresponding to one label may also share properties with another category. Therefore, InfoNCE applies an overly rigid constraint, forcing each motion latent to align exactly with a fixed class token.
In contrast, our method employs a soft cosine-similarity loss that only encourages positive text–motion pairs to move closer, while preserving the unique characteristics of each latent.
This softer regularization achieves better quantitative results and leads to a more semantically meaningful and well-separated latent space. 

\paragraph{Exploration with the latent size}
There are two key factors that determine the size of our motion latent space: (1) the number of latents, and (2) the dimensionality of each latent. \figref{fig:latent_dim} presents a comprehensive ablation over both aspects. We observe that increasing the number of latents, i.e., moving from 4× to 2× compression, leads to consistently better R-Precision scores, indicating stronger text–motion alignment due to the preservation of more fine-grained motion details. In contrast, increasing the latent dimensionality has a negative effect on overall performance.


\subsection{Comparison with SOTA Methods}
\label{sec:exp_comp}
We finally select the model using auto-regressive generation with T5 + CrossAttn conditioning, at 4× temporal downsampling and a 16-d latent space, as our best configuration. We choose two variants with VAE and SAE respectively. In this section, we present qualitative comparisons, quantitative results, and a user study.

\paragraph{Qualitative comparison}
~\cref{fig:gallery} shows qualitative comparisons with MARDM~\cite{meng2024rethinking_marmdm}, ACMDM~\cite{meng2025absolute} and MotionStreamer~\cite{xiao2025motionstreamer}. Our method generates more realistic motions, even for challenging cases like cartwheels and elbow touching knees. As static figures cannot fully capture motion differences, we recommend viewing the supplementary videos for a better comparison.

\paragraph{Quantitative comparison using MARDM-67 protocol}
We present the comparison in \tabref{tab:comp_guo_67}. Our approach outperforms prior methods in both distributional quality and text–motion faithfulness, reducing FID from $0.053$ to $0.049$ for our VAE variant and improving R-Precision (Top-1) from $0.522$ to $0.542$ for our SAE variant. All results are computed on the HumanML3D test set over 20 generation runs, and we report the mean along with the $95\%$ confidence intervals, ensuring the robustness, motion realism, and semantic faithfulness of our method. For results under other evaluation protocols, we refer readers to the supp. material.


\paragraph{User study}
We conducted a user study to evaluate the perceptual quality and text–motion alignment of our generated results. We compare our method against three prior works: the VQ-based methods MoMask~\cite{guo2023momask} and DisCoRD~\cite{cho2024discord}, and the auto-regressive continuous-space model MotionStreamer~\cite{xiao2025motionstreamer}. We randomly selected 20 pairs of sequences per baseline. To reduce potential bias, the order of the placement of methods were randomized.
Participants were instructed to evaluate the realism and text-motion alignment of the generated motions. For each comparison, we collected responses from 15 users. Please refer to the supplementary material for details of the user study interface.
Participants preferred our method over DisCoRD in $83.75\%$ of cases, over MoMask~\cite{guo2023momask} in $77.70\%$ of cases, and over MotionStreamer in $84.70\%$ of cases.
These findings indicate that users consistently perceived our generated motions as more realistic than those of prior works. We generate our results using a $272$D representation when comparing with MotionStreamer, and a $263$D representation when comparing with the other methods, ensuring there are no confounding factors arising from the choice of motion representation.

\begin{figure}[t]
\vspace{-3mm}
\centering
    \begin{overpic}[trim=0cm 0cm 0cm 0cm,clip, width=\linewidth]{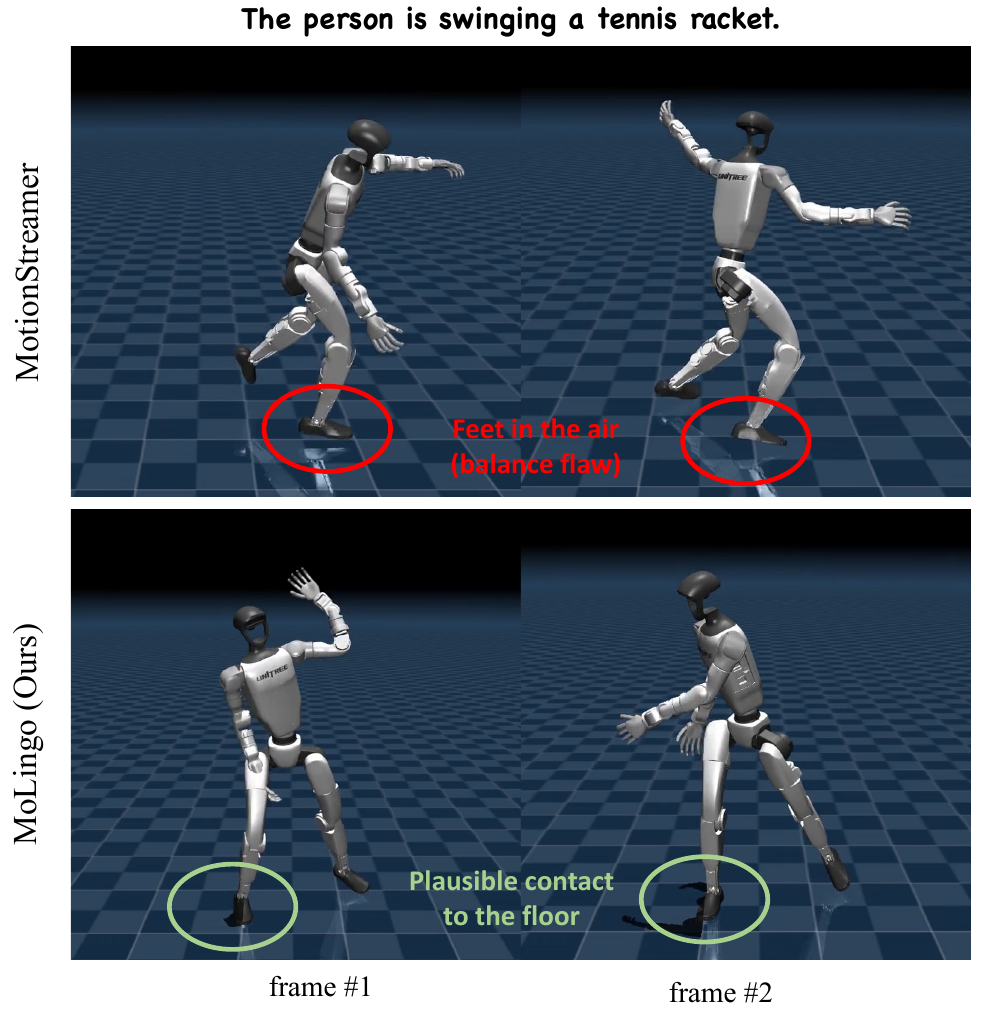}
    \end{overpic}
    \caption{\textbf{Incorporating MoLingo with RL Tracking Controller} Our method produces motion with consistently realistic foot–ground interaction across the entire sequence. In contrast, MotionStreamer frequently exhibits balance artifacts. By leveraging MoLingo within the RL tracking controller, our results maintain stable, physically plausible floor contact, leading to more balanced, grounded, and visually convincing character motion. This improved contact behavior also reduces foot sliding and penetration, highlighting the advantage of our approach in capturing physically coherent movement.
    }
    \vspace{-5mm}
	\label{fig:robot}
 \vspace{-0.5pt}
\end{figure}

\paragraph{Transfer to human robots} A main application generative human motion models is to transfer motion to robots~\cite{li2025interagent, tevet2025closd, wu2025uniphys}. To demonstrate the realism of our method, we show that we can directly use a pre-trained controller with physical constraints to transfer our generations to physics simulation. We compare our method to the MotionStreamer on these tasks see \figref{fig:robot}. The qualitative and quantitative superiority of our method is also reflected in this task.
\vspace{-2mm}\section{Conclusion and Limitation}
\label{sec:conclusion}\vspace{-2mm}

In this work, we present \method{}, a text-to-motion generation model that performs masked auto-regressive rectified flow in a continuous motion latent space aligned with text. By exploring a more structured latent space and a better text condition mechanism, our method produces natural, smooth motions that better follow the input descriptions and achieves state-of-the-art results on a series of evaluators in terms of FID, R-Precision, and CLIP-based scores. A key finding is that aligning the motion latents with text makes the latent space much more aligned with the semantic structure of text models such improves text–motion alignment a lot while preserving good motion quality. We also showed that using multiple text tokens with cross-attention gives stronger text conditioning than a single token, leading to more accurate and realistic generated motions.

\paragraph{Limitations} Since our focus is on main body dynamics, our approach does not generate full-body motion with detailed hand movements, an essential aspect of human motion. We see this as an exciting future direction. 

\newpage
{\small
\paragraph{Acknowledgments:} 
We thank Chuqiao Li and István Sárándi for their helpful discussions and proofreading. We also thank Junyu Zhang for valuable discussions regarding the G1 policy training.
The project was made possible by funding from the Carl Zeiss Foundation. 
This work is supported by the Deutsche Forschungsgemeinschaft (DFG, German Research Foundation) - 409792180 (Emmy Noether Programme, project:  Real Virtual Humans) and the German Federal Ministry of Education and Research (BMBF): Tübingen AI Center, FKZ: 01IS18039A. 
JEL is supported by the German Research Foundation (DFG) - 556415750 (Emmy Noether Programme, project: Spatial Modeling and Reasoning).
GPM is a member of the Machine Learning Cluster of Excellence, EXC number 2064/1 - Project number 390727645. This work was supported by the Engineering and Physical Sciences Research Council [grant number EP/X011364/1]. P. B. is supported by the Konrad Zuse School of Excellence in Learning and Intelligent Systems (ELIZA) through the DAAD programme Konrad Zuse Schools of Excellence in Artificial Intelligence, sponsored by the Federal Ministry of Education and Research. T. B. was supported by a UKRI Future Leaders Fellowship (MR/Y018818/1) as well as a Royal Society Research Grant (RG/R1/241402).
}

{
    \small
    \bibliographystyle{ieeenat_fullname}
    \bibliography{main}
}

\clearpage
\setcounter{page}{1}
\maketitlesupplementary

\begin{table}[h!]
    \centering
    \caption{\label{tab:sym_table}\textbf{Notation Table}. The main notation used in our paper.}
    \small
    \setlength{\tabcolsep}{3pt}
    \begin{tabular}{lll}
    \toprule
    Symbol               & Description                          & Domain \\
    \hline
    $ N $          & Input motion sequence length          & $\mathbb{N}$ \\
    $ D $          & Pose configuration dimension          & $\mathbb{N}$ \\
    $ l $          & Motion latent length                  & $\mathbb{N}$ \\
    $ d $          & Latent dimension                     & $\mathbb{N}$ \\
    $ D_{h} $          & Hidden size of Transformer decoder                     & $\mathbb{N}$ \\
    $ l_{\text{adapter}} $          & Text adapter depth   & $\mathbb{N}$ \\
    $ l_{\text{text}} $          & Text token length for cross attention    & $\mathbb{N}$ \\
    $ \tau $          & Threshold for filtering repetitive class tokens    & $\mathbb{N}$ \\
    \hline
    $ \mathbf{m} $         & Input motion sequence     & $\mathbb{R}^{N\times D}$ \\
    $ m $         & Motion latent sequence     & $\mathbb{R}^{l\times d}$ \\
    $ \kappa $         & Class token sequence     & $\mathbb{R}^{l\times d}$ \\
    $ \mathcal{I} $         & Index set of chosen positions for $\mathcal{L}_{\text{sem}}$  & -  \\
    $ \mathbf{w} $         & Text representation for cross attention     & $\mathbb{R}^{l_{\text{text}}\times D_{h}}$ \\
    \hline
    $ \mathcal{E} $ & Encoder & - \\
    $ \mathcal{D} $ & Decoder & - \\
    $ v_{\theta} $ & Rectified flow MLP & - \\
    
    \bottomrule
    \end{tabular}
    \label{tab:notation}
    \vspace{-10pt}
\end{table}

In this supplementary document, we first present additional implementation details, including autoencoder training, the auto-regressive denoising model, integration with the RL tracking controller, and the evaluation metrics. Next, we report further quantitative results, such as effect of text adapter, comparisons with MotionStreamer~\cite{xiao2025motionstreamer} and a broader set of benchmarks under the TMR-$263$~\cite{petrovich23tmr} evaluation protocol. Finally, we include an ablation study that highlights the impact of our repetitive class-token filtering design during SAE training. The main notation used in our paper is shown in \tabref{tab:notation}.


\begin{table}[t]
\centering
\caption{\textbf{Effect of text adapter depth.} We compare models without a text adapter and with text adapters of different depths ($3$, $6$, and $9$ layers). Using a $6$-layer adapter (ours) gives the best overall trade-off, improving FID and R-Precision across all others. We conduct all experiments on MARDM-$67$~\cite{meng2024rethinking_marmdm} evaluator using a VAE with a downsampling 4× and latent dimension $16$.}
\resizebox{\columnwidth}{!}{
\begin{tabular}{lcccc}
\toprule
\multirow{2}{*}{Configuration} &
\multirow{2}{*}{FID\,$\downarrow$} &
\multicolumn{3}{c}{R-Precision\,$\uparrow$}  \\
\cline{3-5}
 & & Top 1\, & Top 2\, & Top 3\ \\
\midrule
w/o text adapter & $0.057^{\pm{.001}}$ & $0.524^{\pm{.003}}$ & $0.715^{\pm{.001}}$ & $0.810^{\pm{.002}}$  \\ 
depth=$3$ & $0.060^{\pm{.002}}$ & $0.521^{\pm{.002}}$ & $0.713^{\pm{.002}}$ & $0.807^{\pm{.002}}$  \\ 
depth=$6$ (Ours) &\cellcolor{lightgreen}$0.049^{\pm{003}}$ &\cellcolor{lightgreen}$0.528^{\pm{.002}}$ & \cellcolor{lightgreen}$0.721^{\pm{.002}}$ &\cellcolor{lightgreen}$0.815^{\pm{.002}}$  \\
depth=$9$ &\cellcolor{lightyellow}$0.053^{\pm{.002}}$ &\cellcolor{lightyellow}$0.527^{\pm{.002}}$ & \cellcolor{lightyellow}$0.719^{\pm{.002}}$ & \cellcolor{lightyellow}$0.813^{\pm{.001}}$  \\
\bottomrule
\end{tabular}}
\label{tab:abl_adapter}
\end{table}


\begin{table}[t]
\centering
\caption{\textbf{Effect of repetitive class token filtering.} We ablate generative performance with and without repetitive class token filtering during SAE training. Filtering consistently improves FID and retrieval scores, indicating that it acts as a soft semantic regularizer, encouraging coherence without forcing adjacent motion latents to collapse to the same text label.}
\resizebox{\columnwidth}{!}{
\begin{tabular}{lcccc}
\toprule
\multirow{2}{*}{Configuration} &
\multirow{2}{*}{FID\,$\downarrow$} &
\multicolumn{3}{c}{R-Precision\,$\uparrow$}  \\
\cline{3-5}
 & & Top 1\, & Top 2\, & Top 3\ \\
\midrule
w/o filtering & \cellcolor{lightyellow}$0.096^{\pm{.003}}$ & \cellcolor{lightyellow}$0.524^{\pm{.002}}$ & \cellcolor{lightyellow}$0.713^{\pm{.002}}$ & \cellcolor{lightyellow}$0.811^{\pm{.002}}$  \\ 
w/ filtering (Ours) &\cellcolor{lightgreen}$0.066^{\pm{003}}$ &\cellcolor{lightgreen}$0.544^{\pm{.002}}$ & \cellcolor{lightgreen}$0.739^{\pm{.002}}$ &\cellcolor{lightgreen}$0.832^{\pm{.002}}$  \\
\bottomrule
\end{tabular}}
\vspace{-0.2cm}
\label{tab:filter}
\vspace{-0.3cm}
\end{table}

\section{More Implementation Details}
\label{sec:more_details}
\paragraph{Autoencoder training} 
We adopt the causal autoencoder architecture from~\cite{xiao2025motionstreamer} with a hidden size of $1024$, and provide a detailed specification in \tabref{tab:ae_arch}.
Unlike~\cite{xiao2025motionstreamer}, we address an issue happening when the 1D convolutional kernel slides from the beginning to the end of the motion sequence: kernels at the start of the sequence encounter padded values (set to $0$ in previous works), which leads to jitter in the decoded motion during the first few frames. To mitigate this, for the first convolutional layer of the encoder, we replace zero padding with replicated padding, resulting in more stable training and smoother reconstructions.
\tabref{tab:comp_tmr_272} shows our SAE achieves lower reconstruction FID (rFID) while maintaining comparable MPJPE compared with MotionStreamer~\cite{xiao2025motionstreamer}'s VAE.
In practice, we set $\lambda_{\text{joint}} = 1.0$ and $\lambda_{\text{vel}} = 10.0$. For both the VAE and SAE, we use $\lambda_{\text{KL}} = 1\times 10^{-5}$. For the SAE, based on the ablation results in \tabref{tab:abl_sae}, we set $\lambda_{\text{sem}} = 0.001$.
The threshold $\tau$ for filtering repetitive class token is set to $0.995$.

\paragraph{Obtaining text labels for a specific motion latent} 
During SAE training, to obtain text labels that are temporally aligned with a given motion latent $m_i$, we exploit the causal design: each latent depends only on past frames, not future ones. As shown in \tabref{tab:ae_arch}, our encoder comprises four causal 1D convolutional layers in total (excluding the ResNet layers). The first and last layers use a kernel size of $3$ and stride of $1$ to preserve the sequence length, while the two middle layers use a kernel size of $4$ and stride of $2$, each downsampling the temporal resolution by a factor of $2$. This yields a latent sequence whose length is one quarter of the original.
As a result, latent index $i$ is influenced by the original frames from $4i-16$ to $4i+4$, covering $20$ frames in total. We aggregate the frame-level text labels over the same temporal window from $4i-16$ to $4i+4$, encode them with T5-Large to obtain embeddings, and then take the average of these embeddings before feeding them into an MLP projection.

\paragraph{Auto-regressive motion latent denoising}
Our Transformer for predicting $z$ is a standard Transformer decoder~\cite{vaswani2017attention} with $16$ layers, each using $16$ attention heads and a hidden size of $D_h=1024$.
For denoising, we use an MLP with $8$ residual blocks and a width of $1280$ channels. Each block applies LayerNorm, a linear layer, a SiLU~\cite{elfwing2018sigmoid} activation, and another linear layer, followed by a residual connection. The MLP is conditioned on the $z$ predicted by the Transformer: we add $z$ to the time embedding of the noise-schedule step $t$, and use this combined signal to modulate the LayerNorm layers via AdaLN. We set $l_{\text{text}}=128$.
During inference, we use a stochastic sampler that, at each step, applies a small noise refresh, partially replacing the clean component with fresh prior noise. This maintains stochasticity and improves sample diversity compared to a purely deterministic ODE trajectory. To avoid confusion, we distinguish two terms.  \textbf{Denoising step} is the number of denoising iterations for the reverse ODE process during denoising, whereas \textbf{inference step} is the number of auto-regressive iterations used to sample the motion latent sequence. 
\tabref{tab:speed} shows MoLingo can generate per motion \textbf{under 1 sec.}  (e.g., \textbf{0.83s} ) while still maintaining strong generative quality, outperforming the majority of baselines. Meanwhile, higher steps consistently result in lower FID and higher R-Precision, so after tuning this trade-off we adopt \textbf{16 inf. steps and 32 samp. steps} as our default setting: it delivers SOTA performance while keeping generation faster compared to higher-step settings.

\begin{table}[t]
\centering
\caption{\textbf{Averaged Inference Time} (AIT, in seconds) 
and generative performance under different inference and sampling steps.
}
\vspace{-4mm}
\resizebox{\columnwidth}{!}{
\begin{tabular}{c|c|ccccccc}
\toprule
Inf. & Samp. &
\multirow{2}{*}{AIT(s)\,$\downarrow$} &
\multirow{2}{*}{FID\,$\downarrow$} &
\multicolumn{3}{c}{R-Precision} &
\multirow{2}{*}{CLIP-Score\,$\uparrow$} \\
\cline{5-7}
Step & Step & & & Top 1$\uparrow$ & Top 2$\uparrow$ & Top 3$\uparrow$ & \\
 \hline
 $49$ & $32$ & $9.02$ & $\cellcolor{lightgreen}0.056^{\pm{.005}}$ & \cellcolor{lightyellow}$0.543^{\pm{.003}}$ & \cellcolor{lightyellow}$0.737^{\pm{.002}}$ & $0.830^{\pm{.002}}$ & \cellcolor{lightyellow}$0.685^{\pm{.001}}$  \\
$24$ & $32$  & $4.43$ & \cellcolor{lightyellow}$0.057^{\pm{.004}}$ & $0.541^{\pm{.002}}$ & \cellcolor{lightyellow}$0.737^{\pm{.002}}$ & \cellcolor{lightgreen}$0.833^{\pm{.002}}$ & \cellcolor{lightyellow}$0.685^{\pm{.001}}$  \\
$16$ & $32$  & $2.83$ & $0.066^{\pm{.003}}$ & \cellcolor{lightgreen}$0.544^{\pm{.002}}$ & \cellcolor{lightgreen}$0.739^{\pm{.002}}$ & \cellcolor{lightyellow}$0.832^{\pm{.002}}$ & \cellcolor{lightgreen}$0.686^{\pm{.001}}$  \\
$5$ & $32$  & $0.83$ & $0.105^{\pm.005}$ & $0.537^{\pm.003}$ & $0.734^{\pm.002}$ & $0.829^{\pm.002}$ & \cellcolor{lightyellow}$0.685^{\pm.001}$ \\
\hline
$16$ & $8$  & $1.58$ & $0.066^{\pm{.003}}$ & $0.541^{\pm{.002}}$ & \cellcolor{lightyellow}$0.737^{\pm{.002}}$ & $0.830^{\pm{.002}}$ & $0.683^{\pm{.001}}$  \\
$16$ & $100$  & $7.00$ & $0.067^{\pm{.002}}$ & \cellcolor{lightyellow}$0.543^{\pm{.002}}$ & \cellcolor{lightgreen}$0.739^{\pm{.002}}$ & \cellcolor{lightgreen}$0.833^{\pm{.002}}$ & \cellcolor{lightyellow}$0.685^{\pm{.001}}$  \\
\bottomrule
\end{tabular}}
\label{tab:speed}
\end{table}

\begin{figure}[t]
\vspace{-3mm}
\centering
    \begin{overpic}[trim=0cm 0cm 0cm 0cm,clip, width=\linewidth]{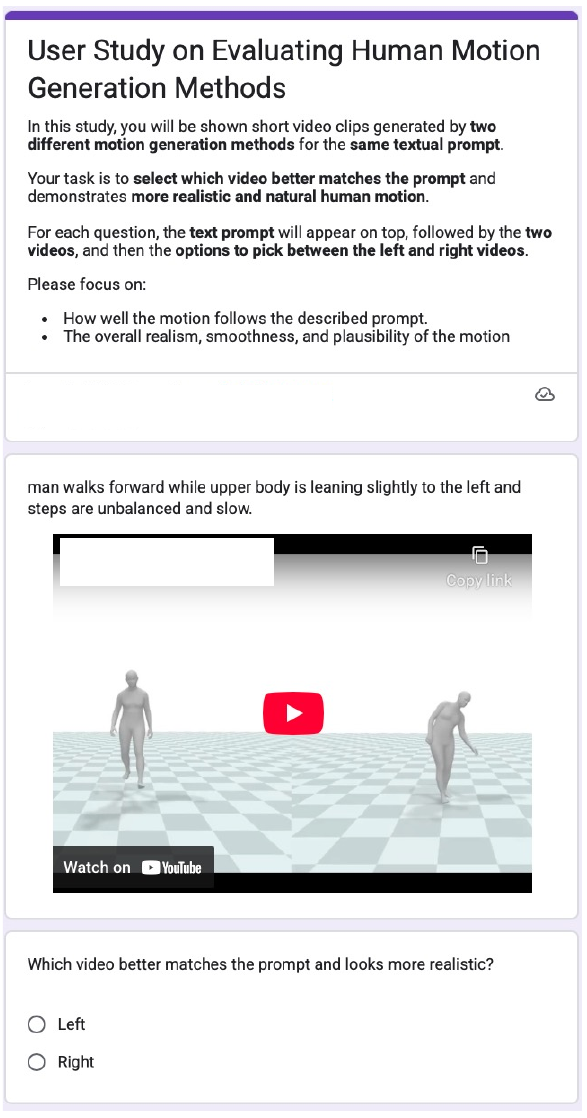}
    \end{overpic}
    \caption{\textbf{User study interface.} }
    \vspace{-3mm}
	\label{fig:user_study}
 \vspace{-0.5pt}
\end{figure}

\paragraph{User study interface}
The detailed instructions and interface layout of our user study are shown in \figref{fig:user_study}. In each trial, we randomly select $15$ samples and randomly assign them to the left or right side.
We prepare three questionnaires with the same layout, each comparing our method against a different baseline.
For comparison with MotionStreamer~\cite{xiao2025motionstreamer}, we use the model trained with the $272$D representation, and for comparisons with MoMask~\cite{guo2023momask} and DisCoRD~\cite{cho2024discord}, we use the model trained with the $263$D representation.


\paragraph{Integrate MoLingo with RL tracking controller}
To make the generated motions more physically plausible, we integrate each generator with a pre-trained RL tracking controller.
Specifically, we first re-train a PHC~\cite{luo2023perpetual} policy for the Unitree G1 humanoid to track the joint positions of human motions. After generating motions with each candidate generator, we retarget them to the humanoid’s body shape and then directly deploy the tracking policy to assess performance.
Because the $272$D representation contains valid rotational components, retargeting is more convenient; thus, we conduct this evaluation using the $272$D model and compare against MotionStreamer.
The policy is trained in IsaacGym~\cite{makoviychuk2021isaac} with $2048$ parallel environments.
After obtaining the generated motions, we first retarget them to match G1’s morphology and extract the corresponding joint position sequences. These joint positions are then provided to the policy to roll out action trajectories. Finally, we replay the resulting actions in the MuJoCo~\cite{todorov2012mujoco} for visualization. Video comparisons are available on our local webpage.


\begin{table*}[t]
\begin{center}
\centering
\caption{\textbf{Quantitative comparison with MotionStreamer.}
MotionStreamer proposed a TMR-style~\cite{petrovich23tmr} feature extractor used as an evaluator in their own $272$D representation. To ensure a fair comparison, we train our model on the HumanML3D-$272$~\cite{xiao2025motionstreamer} dataset and evaluate using their evaluator. 
rFID denotes reconstruction FID, and MPJPE is measured in millimeters. Our replicated padding design improves reconstruction realism in terms of rFID and MPJPE.
For generation, our method achieves significant improvements over MotionStreamer across all metrics.
We run the experiment $20$ runs and $\pm$ indicates $95\%$ confidence interval. We apply $2\times$ downsampling and latent dimension $32$ with a well-trained SAE.
}
\resizebox{0.95\linewidth}{!}{
\begin{tabular}{l|cc|cccccc}
\toprule
\multirow{2}{*}{Methods} &
\multirow{2}{*}{rFID\,$\downarrow$} &
\multirow{2}{*}{MPJPE\,$\downarrow$} &
\multirow{2}{*}{FID\,$\downarrow$} &
\multicolumn{3}{c}{R-Precision} &
\multirow{2}{*}{Matching Score\,$\downarrow$} \\
\cline{5-7}
 & & & & Top 1$\uparrow$ & Top 2$\uparrow$ & Top 3$\uparrow$ & & \\
\midrule
Real & $0.000 $ & $0.000$ & $0.002^{\pm.000}$ & $0.702^{\pm.000}$ & $0.864^{\pm.000}$ & $0.914^{\pm.000}$ & $15.151^{\pm.000}$  \\
\hline
MotionStreamer~\cite{xiao2025motionstreamer} & \cellcolor{lightyellow}$0.661 $ & \cellcolor{lightyellow}$22.90$ & \cellcolor{lightyellow}$11.979^{\pm.078}$ & \cellcolor{lightyellow}$0.629^{\pm.004}$ & \cellcolor{lightyellow}$0.791^{\pm.003}$ & \cellcolor{lightyellow}$0.858^{\pm.003}$ & \cellcolor{lightyellow}$16.019^{\pm.017}$   \\
MoLingo (Ours)  & \cellcolor{lightgreen}$0.026 $ & \cellcolor{lightgreen}$9.87$  & \cellcolor{lightgreen}$3.415^{\pm.034}$ & \cellcolor{lightgreen}$0.776^{\pm.003}$ & \cellcolor{lightgreen}$0.902^{\pm.002}$ & \cellcolor{lightgreen}$0.941^{\pm.001}$ & \cellcolor{lightgreen}$14.712^{\pm.017}$\\
\bottomrule
\end{tabular}}
\label{tab:comp_tmr_272}
\end{center}
\end{table*}


\begin{table*}[th]
\begin{center}
\centering
\caption{\textbf{Quantitative results on the TMR-263 evaluator.} We compare our method with a broad set of motion generation approaches, from early models~\cite{tevet2023human, chen2023executing} to recent ones~\cite{cho2024discord, zhu2025motiongpt3}, covering pose-frame diffusion~\cite{tevet2023human}, single-vector latent diffusion~\cite{chen2023executing, motionlcm-v2}, VQ-based next-token prediction~\cite{guo2023momask, pinyoanuntapong2024mmm, pinyoanuntapong2024bamm, yuan2024mogents, cho2024discord, guo2025snapmogen}, and continuous-valued auto-regressive models~\cite{zhu2025motiongpt3}. We report the mean results over 20 independent runs, and the $\pm$ values indicate the $95\%$ confidence interval. We do not compare with MARDM~\cite{meng2024rethinking_marmdm} and ACMDM~\cite{meng2025absolute} here because of the motion representation inconsistency. Our method achieves state-of-the-art FID, R-Precision, and MultiModality. Green cells highlight the best scores, and yellow cells the second best.}
\resizebox{0.8\textwidth}{!}{
\begin{tabular}{lccccc}
\toprule
\multirow{2}{*}{Methods} &
\multirow{2}{*}{FID\,$\downarrow$} &
\multicolumn{3}{c}{R-Precision} &
\multirow{2}{*}{MModality\,$\uparrow$} \\
\cline{3-5}
 & & Top 1$\uparrow$ & Top 2$\uparrow$ & Top 3$\uparrow$ & \\
\hline
Real  & $0.000^{\pm.000}$ & $0.676^{\pm.002}$ & $0.810^{\pm.002}$ & $0.861^{\pm.002}$ & - \\
\midrule
MDM-50Step~\cite{tevet2023human} & $0.093^{\pm.000}$ & $0.563^{\pm.007}$ & $0.723^{\pm.007}$ & $0.794^{\pm.006}$ & \cellcolor{lightgreen}$34.207^{\pm.598}$   \\ 
MLD~\cite{chen2023executing} & $0.052^{\pm.000}$ & $0.579^{\pm.006}$ & $0.729^{\pm.007}$ & $0.797^{\pm.004}$ & \cellcolor{lightyellow}$32.481^{\pm.678}$  \\ 
MoMask~\cite{guo2023momask} & $0.022^{\pm.000}$ & $0.687^{\pm.002}$ & $0.825^{\pm.002}$ & $0.877^{\pm.002}$ & $18.406^{\pm.583}$  \\
MMM~\cite{pinyoanuntapong2024mmm} & $0.024^{\pm.000}$ & $0.710^{\pm.002}$ & $0.834^{\pm.002}$ & $0.881^{\pm.001}$ & $15.286^{\pm.577}$  \\
BAMM~\cite{pinyoanuntapong2024bamm}  & $0.044^{\pm.000}$ & $0.652^{\pm.002}$ & $0.790^{\pm.002}$ & $0.845^{\pm.002}$ & $28.703^{\pm.672}$  \\
MogenTS~\cite{yuan2024mogents} & \cellcolor{lightyellow}$0.017^{\pm.000}$  & $0.687^{\pm.003}$ & $0.819^{\pm.002}$ & $0.870^{\pm.002}$ & $12.102^{\pm.394}$  \\
MLD++~\cite{motionlcm-v2} & $0.019^{\pm.000}$  & $0.736^{\pm.003}$ & $0.863^{\pm.002}$ & $0.907^{\pm.002}$ & $26.515^{\pm.621}$\\
MotionLCM-V2~\cite{motionlcm-v2} & $0.026^{\pm.000}$  & $0.751^{\pm.003}$ & \cellcolor{lightyellow}$0.879^{\pm.001}$ & \cellcolor{lightyellow}$0.921^{\pm.001}$  & $28.424^{\pm.753}$  \\ 
DisCoRD~\cite{cho2024discord} & $0.020^{\pm.000}$ & $0.692^{\pm.002}$ & $0.828^{\pm.002}$ & $0.878^{\pm.001}$  & $18.804^{\pm.613}$  \\ 
MotionGPT3~\cite{zhu2025motiongpt3} & $0.021^{\pm.000}$ & \cellcolor{lightyellow}$0.758^{\pm.002}$ & $0.873^{\pm.002}$ & $0.915^{\pm.001}$ & $20.821^{\pm.426}$  \\ 
MoMask++~\cite{guo2025snapmogen} & $0.020^{\pm.000}$ & $0.692^{\pm.003}$ & $0.824^{\pm.002}$ & $0.874^{\pm.002}$ & $16.589^{\pm.439}$  \\ 
\midrule
MoLingo  & \cellcolor{lightgreen}$0.015^{\pm.000}$ & \cellcolor{lightgreen}$0.771^{\pm.002}$ & \cellcolor{lightgreen}$0.888^{\pm.001}$ & \cellcolor{lightgreen}$0.925^{\pm.001}$ & $20.400^{\pm.519}$  \\
\bottomrule
\end{tabular}}
\vspace{-0.2cm}
\label{tab:comp_tmr_263}
\vspace{-0.3cm}
\end{center}
\end{table*}

\paragraph{Details of evaluation metrics} We use the evaluation metrics following~\cite{Guo2022CVPR, meng2024rethinking_marmdm, xiao2025motionstreamer}: (1) \textbf{Fréchet Inception Distance (FID)} quantifies how close the generated motions are to real ones by comparing the overall distributions of predicted and ground-truth motion sequences, by applying a pre-trained feature extractor for both motion and text. (2) \textbf{R-Precision} (Top-1/Top-2/Top-3) together with Matching-score evaluates text–motion consistency by checking whether generated motion embeddings retrieve their paired text prompts among the top candidates. (3) \textbf{MultiModality} (MModality) reflects variation under identical prompts, measuring how diverse the produced motion embeddings are when generating multiple motions for the same text. (4) \textbf{CLIP-Score} measures textual faithfulness via CLIP~\cite{radford2021learning}, computed as the cosine similarity between CLIP embeddings of the generated motion and its text description.

\section{More Quantitative Results}

\paragraph{Effect of the text adapter} \tabref{tab:abl_adapter} reports an ablation over different numbers of text adapter layers. Using the adapter improves generative performance in both FID and R-Precision compared to not using it, as it strengthens text–motion communication during training. We select a depth of $6$ layers, which yields the best performance.

\paragraph{Comparison with MotionStreamer}
To ensure a fair comparison, we use the evaluator from~\cite{xiao2025motionstreamer}, trained on the HumanML3D-$272$ ~\cite{xiao2025motionstreamer} with the TMR contrastive strategy.
We compare FID, R-Precision, and Matching Score as reported under MotionStreamer's original evaluation setting.
As shown in \tabref{tab:comp_tmr_272}, Molingo achieves significant improvements on all metrics. We apply CFG scale $7.0$ for $272$D format generation.
The experiment is conducted with HumanML3D-$272$ test set.

\paragraph{Quantitative results on the TMR-263 evaluator~\cite{petrovich23tmr}}
We further provide, for the first time, a benchmark evaluating a broad set of methods on the TMR-$263$ evaluator. TMR is trained with an improved contrastive loss, yielding a better cross-modal embedding space. We rerun a wide range of baseline methods on this evaluator, as shown in \tabref{tab:comp_tmr_263}, MoLingo still achieves state-of-the-art performance across all methods. The experiment is conducted with HumanML3D~\cite{Guo2022CVPR} test set.

\paragraph{Effect of repetitive token filtering}
During SAE training, to ablate the effect of repetitive class-token filtering, \tabref{tab:filter} reports the generative performance with and without filtering. Removing adjacent repetitive class tokens acts as a softer regularizer and yields significantly better performance.


\begin{table*}
    \centering\setlength{\tabcolsep}{6pt}
    \caption{\textbf{Detail architecture} of our autoencoders. Different from~\cite{xiao2025motionstreamer}, we set the padding mode of the first 1D convolutional layer to replicate the initial frames, which stabilizes training and improves reconstruction.}
    \begin{tabular}{ll}
    \toprule
        Components & Architecture \\ \midrule
        Encoder & (0): CausalConv1D($D$, 1024, kernel\_size=(3,), stride=(1,), dilation=(1,), padding=((2,), mode=$\operatorname{replicate}$)) \\
        ~ & (1): SiLU() \\
        ~ & (2): 2 $\times$ Sequential( \\
        ~ &   ~~~~(0): CausalConv1D(1024, 1024, kernel\_size=(4,), stride=(2,), dilation=(1,), padding=((2,), mode=$\operatorname{zero}$)) \\
        ~ &   ~~~~(1): CausalResnet1D( \\
        ~ &   ~~~~~~~~    (0): CausalResConv1DBlock( \\
        ~ &   ~~~~~~~~~~~~      (activation1): SiLU() \\
        ~ &   ~~~~~~~~~~~~      (conv1): CausalConv1D(1024, 1024, kernel\_size=(3,), stride=(1,), dilation=(9,), padding=((18,), mode=$\operatorname{zero}$)) \\
        ~ &   ~~~~~~~~~~~~      (activation2): SiLU() \\
        ~ &   ~~~~~~~~~~~~      (conv2): CausalConv1D(1024, 1024, kernel\_size=(1,), stride=(1,), dilation=(1,), padding=((0,), mode=$\operatorname{zero}$))) \\
        ~ &   ~~~~~~~~    (1): CausalResConv1DBlock( \\
        ~ &   ~~~~~~~~~~~~      (activation1): SiLU() \\
        ~ &   ~~~~~~~~~~~~      (conv1): CausalConv1D(1024, 1024, kernel\_size=(3,), stride=(1,), dilation=(3,), padding=((6,), mode=$\operatorname{zero}$)) \\
        ~ &   ~~~~~~~~~~~~      (activation2): SiLU() \\
        ~ &   ~~~~~~~~~~~~      (conv2): CausalConv1D(1024, 1024, kernel\_size=(1,), stride=(1,), dilation=(1,), padding=((0,), mode=$\operatorname{zero}$))) \\
        ~ &   ~~~~~~~~    (2): CausalResConv1DBlock( \\
        ~ &   ~~~~~~~~~~~~      (activation1): SiLU() \\
        ~ &   ~~~~~~~~~~~~      (conv1): CausalConv1D(1024, 1024, kernel\_size=(3,), stride=(1,), dilation=(1,), padding=((2,), mode=$\operatorname{zero}$)) \\
        ~ &   ~~~~~~~~~~~~      (activation2): SiLU() \\
        ~ &   ~~~~~~~~~~~~      (conv2): CausalConv1D(1024, 1024, kernel\_size=(1,), stride=(1,), dilation=(1,), padding=((0,), mode=$\operatorname{zero}$))))) \\
        ~ & (3): CausalConv1D(1024, 1024, kernel\_size=(3,), stride=(1,), dilation=(1,), padding=((2,), mode=$\operatorname{zero}$)) \\
        \midrule
        Decoder & (0): CausalConv1D($d$, 1024, kernel\_size=(3,), stride=(1,), dilation=(1,), padding=((2,), mode=$\operatorname{zero}$)) \\
        ~ & (1): SiLU() \\
        ~ & (2): 2 $\times$ Sequential( \\
        ~ &   ~~~~(0): CausalResnet1D( \\
        ~ &   ~~~~~~~~    (0): CausalResConv1DBlock( \\
        ~ &   ~~~~~~~~~~~~      (activation1): SiLU() \\
        ~ &   ~~~~~~~~~~~~      (conv1): CausalConv1D(1024, 1024, kernel\_size=(3,), stride=(1,), dilation=(9,), padding=((18,), mode=$\operatorname{zero}$)) \\
        ~ &   ~~~~~~~~~~~~      (activation2): SiLU() \\
        ~ &   ~~~~~~~~~~~~      (conv2): CausalConv1D(1024, 1024, kernel\_size=(1,), stride=(1,), dilation=(1,), padding=((0,), mode=$\operatorname{zero}$))) \\
        ~ &   ~~~~~~~~    (1): CausalResConv1DBlock( \\
        ~ &   ~~~~~~~~~~~~      (activation1): SiLU() \\
        ~ &   ~~~~~~~~~~~~      (conv1): CausalConv1D(1024, 1024, kernel\_size=(3,), stride=(1,), dilation=(3,), padding=((6,), mode=$\operatorname{zero}$)) \\
        ~ &   ~~~~~~~~~~~~      (activation2): SiLU() \\
        ~ &   ~~~~~~~~~~~~      (conv2): CausalConv1D(1024, 1024, kernel\_size=(1,), stride=(1,), dilation=(1,), padding=((0,), mode=$\operatorname{zero}$))) \\
        ~ &   ~~~~~~~~    (2): CausalResConv1DBlock( \\
        ~ &   ~~~~~~~~~~~~      (activation1): SiLU() \\
        ~ &   ~~~~~~~~~~~~      (conv1): CausalConv1D(1024, 1024, kernel\_size=(3,), stride=(1,), dilation=(1,), padding=((2,), mode=$\operatorname{zero}$)) \\
        ~ &   ~~~~~~~~~~~~      (activation2): SiLU() \\
        ~ &   ~~~~~~~~~~~~      (conv2): CausalConv1D(1024, 1024, kernel\_size=(1,), stride=(1,), dilation=(1,), padding=((0,), mode=$\operatorname{zero}$))))) \\
        ~ &   ~~~~(1): Upsample(scale\_factor=2.0, mode=nearest) \\
        ~ &   ~~~~(2): CausalConv1D(1024, 1024, kernel\_size=(3,), stride=(1,), dilation=(1,), padding=((2,),mode=$\operatorname{zero}$)) \\
        ~ & (3): CausalConv1D(1024, 1024, kernel\_size=(3,), stride=(1,), dilation=(1,), padding=((2,),mode=$\operatorname{zero}$)) \\
        ~ & (4): SiLU() \\
        ~ & (5): CausalConv1D(1024, $D$, kernel\_size=(3,), stride=(1,), dilation=(1,), padding=((2,),mode=$\operatorname{zero}$)) \\
         \midrule
          Autoencoder & (0): Encoder($D$, 1024) \\
          & (1): MLP(1024, $d$) \\
          & (2): Decoder($d$, $D$) \\
        \bottomrule
    \end{tabular}
    \label{tab:ae_arch}
\end{table*}

\end{document}